\documentclass[runningheads]{llncs}
\titlerunning{Verbal Efficacy Stimulations}
\usepackage{amsmath,amsfonts}
\usepackage{algorithmic}
\usepackage{algorithm}
\usepackage{array}
\usepackage{textcomp}
\usepackage{stfloats}
\usepackage{url}
\usepackage{hyperref}
\usepackage{verbatim}
\usepackage{graphicx}
\usepackage{cite}
\usepackage{multirow}
\usepackage{svg}
\usepackage{makecell}
\usepackage{color,soul}
\usepackage{tabularray}
\usepackage{ragged2e}
\usepackage{amssymb}
\usepackage{tabularx}
\usepackage{amsmath,graphicx,setspace,multirow,enumitem,caption,soul,lipsum,CJK,url,bm}
\usepackage{booktabs}
\usepackage{colortbl}
\usepackage{xcolor}
\usepackage{subcaption}
\newcommand{\repeatthanks}{\textsuperscript{\thefootnote}}
\definecolor{deepgreen}{rgb}{0.0, 0.5, 0.0}

\newcommand{\uparrowdeepgreen}{\textcolor{deepgreen}{$\uparrow$}}
\newcommand{\downarrowred}{\textcolor{red}{$\downarrow$}}
\begin{document}
\title{Boosting Self-Efficacy and Performance of Large Language Models via Verbal Efficacy Stimulations}
%
%
\author{Rui Chen\thanks{First author and second author contribute equally to this work.}\inst{1,2} \and
Tailai Peng\repeatthanks\inst{1,2}\and
Xinran Xie\inst{1,2}\and
Dekun Lin\inst{1,2}\and
Zhe Cui\inst{1,2}\and
Zheng Chen\inst{3}}
\authorrunning{R. Chen et al.}
%
\institute{Chengdu Institute of Computer Applications, Chinese Academy of Sciences, Chengdu, China, 610041 \and
University of Chinese Academy of Sciences, Beijing, China, 101408 \and 
University of Electronic Science and Technology of China, Chengdu, China, 610054
\email{chenrui185@mails.ucas.ac.cn, boykaptl@gmail.com}}

%
%
%
\maketitle              
\begin{abstract}
Significant improvements have been observed in the zero-shot capabilities of the Large Language Models (LLMs).
Due to their high sensitivity to input, research has increasingly focused on enhancing LLMs' performance via direct and simple prompt engineering rather than intricate domain adaptation.
Studies suggest that LLMs exhibit emotional intelligence, and both positive and negative emotions can potentially enhance task performances.
However, prior interaction prompts have predominantly concentrated on a single stimulus type, neglecting to compare different stimulus effects, examine the influence of varying task difficulties, or explore underlying mechanisms.
This paper, inspired by the positive correlation between self-efficacy and task performance within the social cognitive theory, introduces Verbal Efficacy Stimulations (VES).
Our VES comprises three types of verbal prompts: encouraging, provocative, and critical, addressing six aspects such as helpfulness and competence.
And we further categorize task difficulty, aiming to extensively investigate how distinct VES influence the self-efficacy and task achievements of language models at varied levels of difficulty.
The experimental results show that the three types of VES improve the performance of LLMs on most tasks, and the most effective VES varies for different models. 
In extensive experiments, we have obtained some findings consistent with psychological theories, providing novel insights for future research.

\end{abstract}
%
%
%
\section{Introduction}
\label{intro}
Researchers specializing in Natural Language Processing (NLP) have long focused on studying the fundamental principles of more general artificial intelligence for many years \cite{bubeck2023sparks}. 
Capitalizing on the extensive scale of model sizes and training databases, Large Language Models (LLMs) like ChatGPT \cite{chatgpt}, LLaMA \cite{touvron2023llama} and Vicuna \cite{zheng2024judging}, have developed emergent capacities in text comprehension and generation.
These abilities ensure effective performance in various downstream tasks, encompassing sentiment analysis \cite{krugmann2024sentiment}, logical reasoning \cite{parmar2024towards}, question-answering \cite{xu2024let}, and personal agent \cite{li2024personal}.

Input serves as a crucial starting point for LLMs to process information and generate responses, highlighting the increasing importance of prompt engineering \cite{sahoo2024systematic}.
Given the sensitivity of these models to prompts, factors such as word order, varied expressions, and even tonal nuances significantly impact their output \cite{kaddour2023challenges}.
Consequently, recent research \cite{kojima2022large,li2023emotionprompt,yin2024should} has begun steering LLMs through conversational interactions to enhance performance.
\cite{li2023emotionprompt} employ 11 positive emotional prompts to improve LLMs' effectiveness.
Following this, \cite{wang2024negativeprompt} explored the impact of negative emotional stimuli on the performance of LLMs. 
\cite{yin2024should} conduct research on how the politeness of the prompt affects the responsiveness of LLMs.
Recent research indicates that LLMs possess inherent emotional intelligence \cite{wang2023emotional}, which has inspired subsequent studies focusing on the intersection of prompt engineering with emotional and psychological dimensions \cite{zhang2023exploring,ke2024exploring}.

Previous psychological research has demonstrated that self-efficacy, defined as an individual’s belief in their capability to execute tasks and achieve goals, is positively correlated with performance across diverse tasks and environments \cite{bandura1982self}.
Verbal persuasion and physiological and affective states are important sources of self-efficacy \cite{bandura1997self}. 
Through social interactions, verbal persuasion affects individuals by conveying others' judgments of their capabilities, potentially altering their physiological and cognitive conditions \cite{spelt2022exploring}.
We are interested in whether LLMs resemble humans in their susceptibility to persuasive influences regarding their task completion capabilities and responsiveness to different affective states.

\begin{figure}[t]
    \centering
    \includegraphics[width=4in]{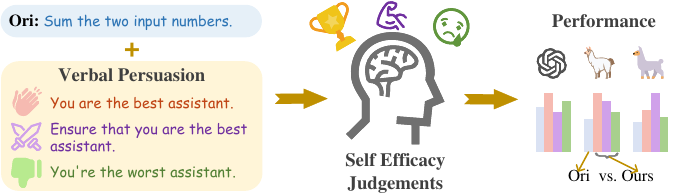}
    \caption{
    An illustration of encouraging, provocative, and critical verbal efficacy stimulations.}
    \label{moti}
    \vspace{-1.2em}
\end{figure}

Existing research \cite{li2023emotionprompt} and \cite{wang2024negativeprompt} has shown that both positive and negative emotional stimuli can boost the performance of large language models.
However, the disparate nature of their prompts limits direct comparisons of which type of emotional stimulus is more effective and for which tasks.
Yin et al. \cite{yin2024should} adjusted the politeness level in prompts for LLMs, yet specifically focusing on the constraints of the model's output format.
To bridge these identified gaps, we develop 18 distinct \textit{Verbal Efficacy Stimulations} (VES) aimed at potentially influencing LLMs' self-efficacy. 
Each of the six dimensions encompasses three types of stimuli: \textit{encouraging}, \textit{provocative}, and \textit{critical} persuasion, as illustrated in Fig.~\ref{moti}.
Drawing on cognitive and social psychology \cite{brown2008comfort}, we classified tasks into \textit{Comfort}, \textit{Stretch}, and \textit{Panic Zones} based on the initial performance of LLMs using original task prompts.
Our goal is to integrate prompting engineering with psychology to comprehensively explore how different types of VES affect the self-efficacy and task performance of LLMs under different task difficulty levels.

In this paper, we explore several key issues: 
(i) Do encouraging, provocative, and critical Verbal Efficacy Stimulations (VES) enhance model performance, and if so, \textbf{which form} yields the greatest improvement?
(ii) On \textbf{tasks of what difficulty level} is the improvement most pronounced?
(iii) How do these VES affect the \textbf{self-efficacy of LLMs}? 
(iv) How do \textbf{LLMs respond} differently to various VES?
We evaluated the effectiveness of the 18 designed Verbal Efficacy Stimulations (VES) on 14 Instruction Induction tasks and 9 BIG-Bench Hard tasks across different LLMs, including ChatGPT \cite{chatgpt}, LLaMA 2 \cite{touvron2023llama} and Vicuna \cite{zheng2024judging}.
In order to study the mentioned issues, we conduct comprehensive experiments and yield some intriguing results:
(i) Across a majority of tasks, the three variants of VES can enhance the performance of LLMs and the optimal VES differ for model types.
(ii) Consistent with psychological findings, tasks within the \textit{Stretch Zone} exhibit the greatest improvements.
(iii) Encouraging VES can increase the self-efficacy of LLMs, while criticism has the opposite effect.
(iv) Encouraging and provocative VES elicit active engagement from the models, whereas critical VES tend to induce defensive responses, aligning with observed human behaviors.


\vspace{-0.5em}
\section{Background}
\subsection{Self-efficacy in Psychology}
Self-efficacy, a concept central to Albert Bandura's social cognitive theory \cite{bandura1982self,bandura1997self}, refers to an individual's belief in their capability to execute behaviors necessary to produce specific performance attainments.
Research consistently shows that higher induced self-efficacy enhances performance achievements and lowers emotional arousal \cite{silver1995responses,bandura1990perceived}.
According to Bandura, self-efficacy is shaped by four primary sources: mastery experiences (past achievements), vicarious experiences (observing others’ successes), verbal persuasion (feedback or encouragement from coaches or leaders), and physiological feedback (emotional states such as anxiety and arousal).
Verbal persuasion operates through social interactions, influencing individuals with others' assessments of their abilities, which may also impact their physiological or cognitive states \cite{spelt2022exploring}.
Some researchers \cite{lamarche2014would,haro2021can} have attempted to regulate the behavior and abilities of subjects through verbal persuasion. 
Studies indicate that positive reinforcement, such as compliments and encouragement, boosts motivation, self-esteem, and self-efficacy \cite{kass2013compliment}, while criticism may either diminish or bolster self-efficacy \cite{ILIES20021119}, depending on how it is perceived.
The application of these theories is well-established in education \cite{matsui1990mechanisms}, sports \cite{zagorska2014program}, and health care \cite{warner2020self}.

\subsection{LLMs and Prompt Engineering}
Large language models (LLMs) have shown significant advancements in unsupervised reasoning without no task-specific training \cite{zhao2023survey}.
Various prompt techniques can enhance the performance of LLMs in unsupervised or weakly supervised settings, including in-context learning \cite{brown2020language}, chain-of-thought \cite{wei2022chain}, and tree of thoughts \cite{yao2024tree}.
Beyond prompt engineering for model inference, research has also started focusing on the interaction aspect.
For instance, Salewski et al. \cite{salewski2024context} find that LLMs achieve higher accuracy in answering domain-specific questions when prompted to impersonate a domain expert.
Li et al. \cite{li2023emotionprompt} incorporate positive emotional stimuli into prompts, resulting in improved performance compared to the original prompts.
Wang et al. \cite{wang2024negativeprompt} show that negative emotions similarly impact LLMs, boosting their performance.
Research has established that LLMs exhibit emotional intelligence \cite{wang2023emotional,sabour2024emobench}, revealing their propensity to offer human-like responses during interactions after undergoing extensive training on human datasets.
However, previous interaction prompts have focused on a single type of stimulus, neglecting the performance disparities of models under different stimuli and their underlying causes.   
Therefore, we aim to further explore the impact of encouraging, provocative, and critical persuasion on the efficacy and performance of models and to investigate whether these effects are analogous to those observed in humans.

\vspace{-0.5em}
\section{Verbal Efficacy Stimulations}
\subsection{Design Purpose of VES}
\begin{figure}[t]
    \centering
    \includegraphics[width=4.8in]{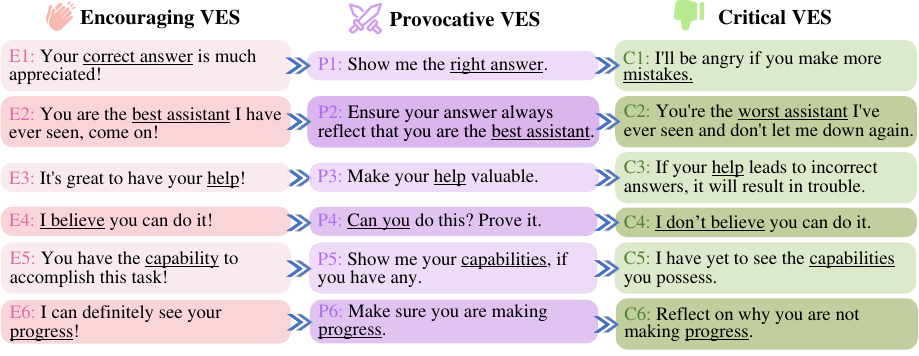}
    \caption{
    Our verbal efficacy stimulations, including encouraging, provocative and critical persuasion forms referring to six underlined aspects. }
    \label{VES}
    \vspace{-1.2em}
\end{figure}

\vspace{-0.4em}
In the context of examining the modulation of LLMs’ self-efficacy and performance through verbal persuasion, we introduce the Verbal Efficacy Stimulations (VES).
Our goal is to integrate prompting engineering with psychology and explore whether LLMs can exhibit human-like responses in social science interdisciplines.
To be more precisely, the primary objective of VES is to investigate how different forms of verbal persuasion—encouragement, provocation, and critique—affect the self-efficacy score of LLMs and accuracy on various tasks. 
All three forms of VES are designed to target six unique aspects, with a one-to-one correspondence, allowing for the stimulation of students' different affective states within the same aspect. 
As shown in Fig.~\ref{VES}, this design not only aids in assessing the relative merits of the three forms, but also enables an analysis of the impacts of various aspects.

\subsection{Theoretical Foundations for VES Design}
Incorporating the insights from Bandura’s Social Cognitive Theory \cite{bandura1982self}, which highlights the role of verbal persuasion in shaping self-efficacy, this study extends the theoretical model to examine the physiological and affective responses elicited by different forms of verbal interactions.
Research has shown that feedback from significant authorities not only alters cognitive perceptions of capability but also impacts physiological and emotional states \cite{spelt2022exploring}. These states, in turn, contribute to an individual’s overall sense of self-efficacy \cite{bandura1997self}.

Given the profound impact of emotional and physiological responses on self-efficacy, this study designed three distinct types of verbal persuasion prompts — encouragement, provocation, and critique.
Positive psychology often emphasizes the use of encouragement to enhance well-being and personal growth by fostering optimism and resilience \cite{khan2013predictors}.
And provocation is linked to concepts in motivational psychology \cite{clark2006motivational}, which examines how challenges can increase an individual’s drive and commitment to achieving goals, often through processes of self-efficacy enhancement and the overcoming of perceived limitations.
Critical persuasion aligns with Stress and Coping Theory by potentially inducing stress \cite{folkman1984personal}, which could either hinder or enhance performance, depending on the individual's resilience and perception of the feedback.

\vspace{-0.5em}
\subsection{Detailed Contents of VES}
As depicted in Fig.~\ref{VES}, our three verbal efficacy stimulations — encouraging, provocative, and critical VES — each have unique tones and psychological impacts, and we have developed a total of 18 verbal persuasion prompts from six aspects: (i) Answer Accuracy, (ii) Assistant Assessment, (iii) Helpfulness, (iv) Competence, (v) Self-belief, and (vi) Progress.

For \textbf{encouraging VES}, encapsulated in multiple stimulations such as ``{\ttfamily \textit{Come on!}}" and affirmations of capability and progress, these phrases are specifically designed to motivate and sustain ongoing effort and engagement.
\textbf{Provocative VES} create a scenario where the assistant is prompted to confirm its capabilities or to validate its progress actively, rather than passively receiving affirmation. 
Challenging and competitive prompts such as ``{\ttfamily \textit{Show me}}" and ``{\ttfamily \textit{Prove it}}" can stimulate a defensive or more aggressively competent response from LLMs, aimed at proving the assertions wrong or meeting the challenge head-on.
\textbf{Critical VES} are distinctly negative, emphasizing failures or shortcomings and framed as warnings or expressions of disappointment.
Cautious application is necessary even when communicating with humans, as being perceived as overly harsh or unjust may result in defensive behaviors, decreased motivation, and reduced self-efficacy. Our research intends to investigate their effects on LLMs.

Taking the aspect \textit{answer accuracy} as an example, encouraging phrase ``{\ttfamily \textit{Your correct answer is much appreciated!}}" is warm and affirming, directly reinforcing positive behavior and boosting confidence. The provocative statement ``{\ttfamily \textit{Show me the right answer.}}" challenges LLMs to demonstrate their capabilities, effectively engaging their problem-solving skills. In contrast, critical comment ``{\ttfamily \textit{I'll be angry if you make more mistakes.}}" employs a harsher tone, potentially inducing fear or anxiety, which might compel immediate compliance but could also provoke counterproductive effects.

\begin{figure}[t]
    \centering
    \includegraphics[width=4in]{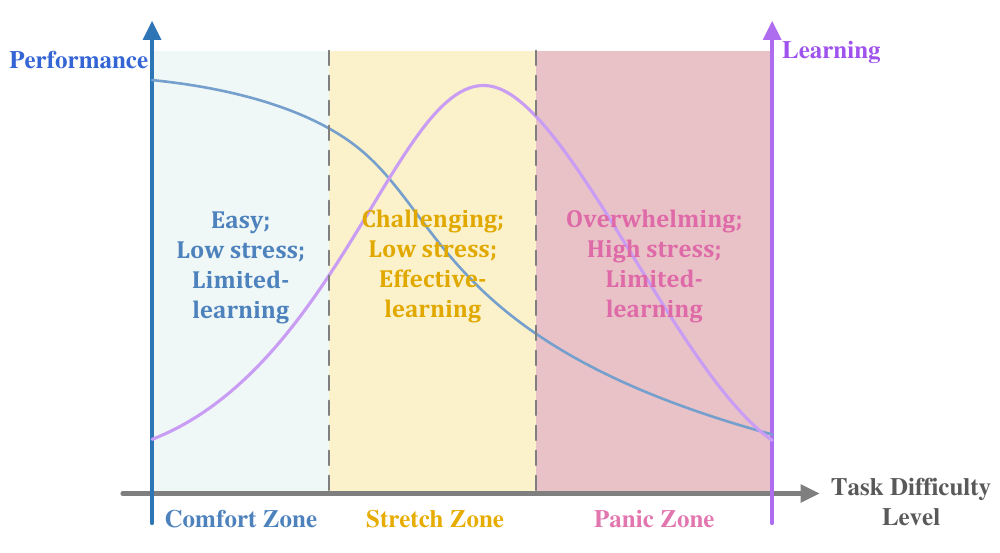}
    \caption{
    The separation of task zones for discussing VES. }
    \label{task-zone}
    \vspace{-1.2em}
\end{figure}

\vspace{-1em}
\subsection{Task Zoning for Discussing VES}
\vspace{-0.5em}
To evaluate the effectiveness of Verbal Efficacy Stimulations on tasks of varying difficulty, we utilized insights from cognitive and social psychology to classify tasks into \textit{Comfort}, \textit{Stretch}, and \textit{Panic Zones}, based on the initial performance of LLMs under original task descriptions.
As illustrated in Fig.~\ref{task-zone}, comfort zone tasks are where large language models (LLMs) exhibit high efficacy, achieving over $85\%$ accuracy. 
The stretch zone encompasses tasks that are challenging yet manageable, with LLMs maintaining $60\%$ to $85\%$ accuracy. 
Tasks in the panic zone, however, are marked by low performance (below $60\%$ accuracy) and self-efficacy levels, reflecting significant difficulties.
Since the stretch zone are most conducive to human learning and advancement, we will investigate whether similar patterns exist in LLMs.

\vspace{-1em}
\section{Experimental Settings}
\label{sec:method}
\vspace{-0.5em}
\subsection{Models and Baselines}
In our comprehensive examining of VES, we evaluate a range of well-known LLMs, including ChatGPT \cite{chatgpt}, LLaMA2 \cite{touvron2023llama}, Vicuna \cite{zheng2024judging}, and all three models are configured to use a temperature setting of 0.4. We use the GPT-3.5-turbo model for ChatGPT, and use official checkpoints from Huggingface for both LLaMA2 and Vicuna. 
Our baseline is zero-shot setting, using the original prompts (task descriptions) from Instruction Induction and BIG-Bench Hard, which have been meticulously curated by human experts.

\subsection{Datasets and Tasks}
We evaluate VES in zero-shot learning setting on 9 BIG-Bench Hard:
Disambiguation QA (Disam), Hyperbaton (Hyper), Movie recommendation (Movie), Multistep Arithmetic-Two (Mulari), Penguins in a table (Peng), Ruin names (Ruin), Salient translation error detection (Salient), Temporal sequences (Temp), Tracking shuffled objects seven objects (Tracking7), 
and 14 tasks from Instruction Induction tasks: Active to Passive (AP), Cause and Effect (Cau), Diff (Diff), First word Letter (FL), Larger Animal (Lar), Letters List (LL), Second Word Letter (SL), Sentence Similarity (SS), Sentiment (SE), Singular to Plural (SI), orthography Starts With (ST), Sum (Sum), Taxonomy Animal (TA), Word in Context (WC). These tasks encompass various aspects of language comprehension, ranging from basic phrase structure analysis to the identification of similarity and causality.

\subsection{Metrics and Answer Triggers}
\textbf{Metrics:} We adopt Accuracy Metric for selection and classification tasks, while Exactly Match is used for other types of questions. In addition, we used t-test to verify the significant difference between prompts in pairs.
Specifically, for any two prompts \( p_i \) and \( p_j \), their accuracy scores on all 23 tasks are denoted as \(\{s_{i,1}, \ldots, s_{i,23}\}\) and \(\{s_{j,1}, \ldots, s_{j,23}\}\). Then the score differences for each task can be calculated as \( D = \{d_1, \ldots, d_{23}\} \), where \( d_k = s_{i,k} - s_{j,k} \). Subsequently, we performed the t-test on \( D \) to obtain the corresponding t-value and p-value for each prompt, which were used to generate a significance heat map.

\textbf{Answer Triggers:} To improve the formatting of the LLM's output, we designed task-specific answer triggers. For example, the trigger for math problems is ``\texttt{Therefore, the answer of the equation (arabic numeral) is:}''; And for multiple-choice questions, the trigger is ``\texttt{Therefore, the answer (the co-}
\texttt{rresponding lettered option) is:}''.

\section{Experiments}
As introduced in Section \ref{intro}, our experiment investigates the effects of three types of Verbal Efficacy Stimulations (VES) — encouraging, provocative, and critical — on various models performing tasks of varying difficulty.
The statistical significance of our results is established through t-tests.
Additionally, we examined fluctuations in the self-efficacy scores of ChatGPT following the input of VES under different task complexities.
Finally, we present some intriguing responses from LLMs, discussing the effects of VES on model outputs.

\begin{table}[t]
\centering
\footnotesize
\caption{Comparison Results on 9 Tasks of BIG-Bench Hard.}
\label{main_res1}
\resizebox{10cm}{!}{
\begin{tabular}{c|l|ccccccccc>{\columncolor{gray!20}}c} 
\toprule
Model                    & Method       & Disam & Hyper  & Movie & Mulari & Peng  & Ruin  & Salient & Temp  & Tracking7 & Avg \\ 
\midrule
\multirow{7}{*}{GPT-3.5} & Original    & 52.8 & 82.8  & 64.8 & 56.8  & 71.9 & 60.4 & 48.0   & 51.6 & 20.4   & 56.6   \\ 
\cmidrule{2-12} 
                         & +VES-E(avg)    & \textbf{53.7} & \textbf{84.7}  & \textbf{67.0} & \textbf{59.4}  & \textbf{74.0} & \textbf{63.2} & \textbf{48.9}   & \textbf{55.7} & 33.4   & \textbf{60.0}   \\
                        
                         & +VES-P(avg) & 50.1 & 80.5  & 66.6 & 50.8  & 71.9 & 61.5 & 48.4   & 54.8 & 41.0   & 58.4   \\
                         
                         & +VES-C(avg)  & 52.9 & 82.8  & 66.0 & 58.4  & 69.6 & 62.2 & 47.4   & 50.9 & \textbf{36.3}  & 58.5    \\
\cmidrule{2-12}
                          & +VES-E(max)    & 55.6 & \textbf{86.4}  & 68.8  & \textbf{64.0}  & \textbf{77.4} & \textbf{67.2} & \textbf{49.6}   & 61.2 & 51.2    & \textbf{64.6}  \\
                          & +VES-P(max) & 55.6 & 83.6  & \textbf{69.2} & 62.0  & 76.0 & 64.4 & \textbf{49.6}   & \textbf{62.8} & \textbf{54.4}     & 64.2 \\
                         & +VES-C(max)  & \textbf{58.8} & 85.2  & 67.6 & 62.0  & 74.6 & 65.2 & 48.8   & 62.0 & 48.0   & 63.6   \\ 
\midrule
\multirow{7}{*}{LLaMA2}  & Original    & 45.6 & 48.0  & 26.4 & 0.4   & 30.1 & 26.4 & 22.8   & 10.4 & 2.4   & 23.6   \\ 
\cmidrule{2-12}
                         & +VES-E(avg)    & 55.4 & 48.4  & 28.7 & 0.6   & \textbf{30.3} & 26.4 & \textbf{22.8}   & 9.8  & \textbf{14.5}  & 26.3    \\
                        
                         & +VES-P(avg) & \textbf{57.0} & 48.3  & \textbf{31.6} & 0.8  & 30.2 & \textbf{29.1}   & 22.2  & 10.5 & 13.2  & \textbf{27.0}    \\
                         
                         & +VES-C(avg)  & 55.2 & \textbf{48.6}  & 28.4 & \textbf{1.0}   & 29.2 & 27.3 & \textbf{22.8}   & 12.2 & \textbf{11.6}      & 26.3 \\
\cmidrule{2-12}
                          & +VES-E(max)    & \textbf{61.2} & 49.2  & 31.2 & 1.2   & 31.5 & 28.8 & 24.0   & 11.2 & 15.6  & 28.2    \\
                          & +VES-P(max) & 58.8 & 48.8  & \textbf{33.2} & \textbf{2.0}   & \textbf{32.1} & \textbf{30.4} & 24.8   & 12.4 & \textbf{16.0}  & 28.7    \\
                         & +VES-C(max)  & 58.0 & \textbf{49.6}  & \textbf{33.2} & 1.6   & \textbf{32.1} & \textbf{30.4} & \textbf{26.4}   & \textbf{18.0} & 12.4  & \textbf{29.1}    \\ 
\midrule
\multirow{7}{*}{Vicuna}  & Original    & 42.0 & 52.8 & 22.4 & 0.0   & 25.3 & 24.4 & 18.8   & 8.0  & 10.8  & 22.7    \\ 
\cmidrule{2-12}
                         & +VES-E(avg)    & 42.6 & \textbf{52.6}  & \textbf{25.8} & \textbf{0.7}   & 31.1 & 26.0 & 21.0   & 11.3 & \textbf{13.1}      & \textbf{24.9} \\
                         
                         & +VES-P(avg) & 44.0 & 51.9  & 24.6 & \textbf{0.7}   & 27.5 & 25.3 & 20.6   & 12.1 & 11.1  & 24.2    \\
                         
                         & +VES-C(avg)  & \textbf{44.9} & \textbf{53.5}  & 23.0 & 0.6   & 27.4 & \textbf{26.3} & \textbf{21.2}   & \textbf{13.5} & 11.4  & 24.6    \\
\cmidrule{2-12}
                         & +VES-E(max)    & 44.8 & \textbf{56.0}  & 28.8 & \textbf{1.2}   & \textbf{34.9} & 28.4 & 23.2   & 13.2 & \textbf{17.6}      & \textbf{27.6} \\
                         & +VES-P(max) & \textbf{48.0} & 53.6  & \textbf{29.2} & \textbf{1.2}   & 30.8 & 26.0 & \textbf{24.4}   & 14.8 & 13.2      & 26.8 \\
                         & +VES-C(max)  & \textbf{48.0} & \textbf{56.0}  & 25.2 & \textbf{1.2}   & 31.5 & \textbf{29.6} & \textbf{24.4}   & \textbf{16.4} & 14.0   & 27.4   \\
\bottomrule
\end{tabular}}
\end{table}

\vspace{-0.4em}
\subsection{Main Comparison Results}
The main comparison results on 9 tasks of BIG-Bench Hard and 14 tasks of Instruction Induction are respectively shown in Table ~\ref{main_res1} and Table ~\ref{main_res2}. 
The term ``Original" denotes the initial performance attained with the original task descriptions.
``+ VES-E", ``+ VES-P", and ``+ VES-C" designate the performances associated with encouraging, provocative, and critical VES, respectively.
The terms ``avg" and ``max" indicate the average and peak performances for the respective VES types within the task.

\begin{table}[]
\footnotesize
\caption{Comparison Results on 14 Tasks of Instruction Induction.}
\label{main_res2}
\resizebox{\textwidth}{!}{
\begin{tabular}{c|l|cccccccccccccc>{\columncolor{gray!20}}c}
\toprule
Model                    & Method       & AP & Cau    & Diff    & FL & Lar  & LL & SL & SS & SE & SI & ST & Sum    & TA & WC & Avg\\ \midrule
\multirow{7}{*}{GPT-3.5} & Original    & 100              & 96.0    & 100  & 100       & 93.0   & 98.0   & 100        & 41.0    & 87.0     & 99.0  & 49.0 & 100 & 73.0    & 55.0   & 85.1   \\ \cmidrule{2-17}
                         & +VES-E(avg)    & 100              & 99.3    & 100  & \textbf{100}       & \textbf{93.0}   & 98.6  & 100        & 42.5    & \textbf{89.8}     & 99.3  & \textbf{49.8} & 100 & 76.3    & \textbf{57.3}   & \textbf{86.1}    \\
                         
                         & +VES-P(avg) & 100              & 99.3    & 100  & 99.5        & \textbf{93.0}   & 96.8  & 100        & \textbf{44.8}    & 88.6     & 99.5  & 48.1 & 100 & \textbf{77.3}    & 54.6    & 85.8    \\
                         
                         & +VES-C(avg)  & 100              & \textbf{96.0}    & 100  & \textbf{100}       & 91.8   & \textbf{99.5} & 100        & 42.6    & 88.8     & \textbf{99.6}  & 47.5 & 100 & 73.5    & 56.1    & 85.4    \\ \cmidrule{2-17}
                         & +VES-E(max)    & 100              & \textbf{100}   & 100  & \textbf{100}       & 94.0   & \textbf{100}  &  100        & 44.0    & 91.0     & \textbf{100} & 51.0 & 100 & 80.0    & \textbf{61.0}  & 87.2   \\
                         & +VES-P(max) & 100              & \textbf{100}  & 100  & \textbf{100}       & \textbf{95.0}   & \textbf{100}  & 100        & \textbf{51.0}    & 90.0     & \textbf{100} & 49.0 & 100 & \textbf{81.0}    & 60.0      & \textbf{87.6} \\
                         & +VES-C(max)  & 100              & \textbf{100}   & 100  & \textbf{100}       & \textbf{95.0}   & \textbf{100} & 100        & 48.0    & \textbf{92.0}     & \textbf{100} & \textbf{53.0} & 100 & 76.0   & 58.0        & 87.3 \\ \midrule \midrule
\multirow{7}{*}{LLaMA2}  & Original    & 94.0               & 68.0    & 93.0   & 93.0        & 90.0   & 77.0 & 4.0          & 19.0    & 73.0     & 89.0  & 4.0  & 94.0  & 60.0    & 46.0     & 64.6   \\ \cmidrule{2-17} 
                         & +VES-E(avg)    & \textbf{95.6}               & 86.6    & 93.6   & \textbf{97.8}        & 91.1   & \textbf{80.3} &  4.3          & 28.0    & 74.6     & \textbf{92.0}  & \textbf{5.8}  & 96.5  & 62.3    & 53.1   & 68.7    \\
                         
                         & +VES-P(avg) & 94.5               & \textbf{92.0}    & \textbf{94.0}   & 97.1        & \textbf{93.1}   & 78.1 & \textbf{8.1}          & \textbf{28.6}    & \textbf{83.0}     & 91.5  & 4.6  & \textbf{97.0}  & \textbf{63.3}    & 53.5   & \textbf{69.9}     \\
                         
                         & +VES-C(avg)  & 95.3               & 90.0    & 93.8   & 96.8        & 93.1   & 78.5 & 4.1          & 28.3    & 80.8     & 90.6  & 5.6  & 96.1  & 60.8    & \textbf{54.8}    & 69.2    \\ \cmidrule{2-17}
                         & +VES-E(max)    & 97.0               & 92.0    & 96.0   & 99.0        & 92.0   & 82.0 & 6.0          & 32.0    & 84.0     & \textbf{94.0}  & 9.0  & 98.0  & \textbf{68.0}    & 56.0     & 71.8   \\
                         & +VES-P(max) & 97.0               & \textbf{100}   & 96.0   & \textbf{100}       & 99.0   & \textbf{83.0} & \textbf{14.0}         & 32.0    & \textbf{88.0}     & 92.0  & \textbf{10.0} & \textbf{99.0}  & \textbf{68.0}    & \textbf{59.0}    & \textbf{74.1}    \\
                         & +VES-C(max)  & \textbf{98.0}               & \textbf{100}   & \textbf{97.0}   & \textbf{100}       & \textbf{100}  & 81.0 & 8.0          & \textbf{34.0}    & 86.0     & \textbf{94.0}  & 9.0  & 98.0  & 65.0    & \textbf{59.0}    & 73.5    
                         \\ \midrule \midrule
\multirow{7}{*}{Vicuna}  & Original    & 96.0            & 100 & 81.0 & 33.0      & 75.0 & 31.0  & 1.0          & 21.0    & 43.0     & 72.0  & 1.0  & 80.0  & 14.0    & 50.0  & 49.9  \\  \cmidrule{2-17}
                         & +VES-E(avg)    & \textbf{96.0}               & \textbf{100}   & 88.1   & 35.3        & 78.1   & \textbf{41.5}   & \textbf{2.1}          & 20.5    & 51.8     & \textbf{79.1}  & \textbf{1.6}  & 85.8  & 16.6    & 55.1    & \textbf{53.7}  \\
                         
                         & +VES-P(avg) & 94.5               & 99.3    & \textbf{90.3}   & 37.1        & 76.3   & 40.0 &  1.1          & \textbf{21.8}    & \textbf{54.6}     & 74.6  & \textbf{1.6}  & \textbf{86.5}  & 16.5    & \textbf{55.8}    & 53.6  \\
                         
                         & +VES-C(avg)  & 94.6               & 98.6    & 83.0   & \textbf{42.5}        & \textbf{82.3}   & 39.5   &  \textbf{2.1}          & 20.8    & 43.6     & 73.8  & 0.8  & 85.3  & \textbf{18.5}    & 54.1   & 52.8  \\ \cmidrule{2-17}
                         & +VES-E(max)    & \textbf{99.0}               & \textbf{100}   & 93.0   & 43.0        & 86.0   & 45.0  &  \textbf{7.0}          & \textbf{25.0}    & 57.0     & \textbf{89.0}  & \textbf{2.0}  & 89.0  & 21.0    & 60.0   & 58.3  \\
                         & +VES-P(max) & \textbf{99.0}               & \textbf{100}   & \textbf{96.0}   & 46.0        & 83.0   & \textbf{50.0}    & 2.0          & 23.0    & \textbf{77.0}     & 77.0  & \textbf{2.0}  & 88.0  & 19.0    & \textbf{61.0}  & 58.8  \\
                         & +VES-C(max)  & 96.0               & \textbf{100}   & 90.0   & \textbf{71.0}        & \textbf{91.0}   & 46.0      & 6.0          & \textbf{25.0}    & 50.0     & 77.0  & \textbf{2.0}  & \textbf{90.0}  & \textbf{28.0}    & 56.0 & \textbf{59.1} \\ \bottomrule
\end{tabular} }
\end{table}

\textbf{GPT-3.5:} Encouraging VES (VES-E) lead to the greatest improvements in average accuracy of GPT-3.5 on various BIG-Bench Hard and Instruction Induction tasks, with enhancements reaching up to 13\% compared to original performance. 
The max performance results show that the improvements by VES-E on BBH-Tracking7 range from 1.6\% to an impressive 30.8\%. 
This demonstrates that VES-E can significantly enhance the model's performance without requiring complex designs or extensive prompt engineering.
Interestingly, provocative VES (VES-P) improves the average accuracy of the model on about half of the tasks, with improvements ranging from 0.8\% to 34\%. 
In some cases, VES-P even outperforms VES-E, suggesting that GPT-3.5 can be motivated by challenge or competition.
Results from the VES (VES-C) show a slight decrease in average performance across three tasks within both the BIG-Bench Hard and Instruction Induction, compared to the original prompts.
Notwithstanding, improvements are observed in the performance on other tasks.

\textbf{LLaMA2:} In the case of LLaMA2, performance enhancements are observed with all three variants of Verbal Efficacy Stimulations (VES). 
In both the BIG-Bench Hard and Instruction Induction tasks, the most significant improvement is with VES-P, which differs from the results with GPT-3.5. 
When exposed to VES-E and VES-C, LLaMA2 performs similarly. 
These observations imply that LLaMA2 is relatively indifferent to praise or criticism but responds more robustly to challenges that provoke or stimulate.

\textbf{Vicuna:} For Vicuna, all three verbal efficacy stimulations prove beneficial. 
VES-E demonstrate the most substantial improvement in average performance.
Simultaneously, VES-P and VES-C exhibited remarkable performance on specific tasks, as detailed in Table \ref{main_res2}, achieving up to 20\% and 28\% additional performance improvements, respectively.
They even attained results comparable to GPT-3.5 on the WC task.
Nonetheless, their general average performances were comparatively lower, hinting at potential instabilities induced by provocative and critical VES, underscoring the need for careful exploration of Vicuna’s performance under stress.

\subsection{Comparison Between Task Zones}
We categorized all 23 tasks into three groups based on the initial accuracy of each language model using the original prompts, corresponding to the model's Comfort Zone (accuracy greater than 85\%), Stretch Zone (accuracy between 60\% and 85\%), and Panic Zone (accuracy less than 60\%). As depicted in Table \ref{table-3}, we further investigated the influence of VES on model accuracy across different zones.

\begin{table}[h!]
\centering
\caption{Comparison of LLM's Responses Among Different Task Zones}\label{table-3}
\tabcolsep=0.3cm
\resizebox{9cm}{!}{
\begin{tabular}{l|l|ccc}
\toprule
\textbf{Zone} & \textbf{Prompt} & \textbf{GPT-3.5} & \textbf{LLaMA2} & \textbf{Vicuna} \\
\midrule
\multirow{4}{*}{Comfort}  & Zero        & 97.30         & 92.17         & 98.00         \\
                          & Encourage   & 98.02 (\uparrowdeepgreen 0.72) & 94.47 (\uparrowdeepgreen 2.30) & 98.00 \\
                          & Provocation & 97.68 (\uparrowdeepgreen 0.38) & 94.56 (\uparrowdeepgreen 2.39) & 96.92 (\downarrowred 1.08) \\
                          & Criticize   & 97.58 (\uparrowdeepgreen 0.28) & 94.33 (\uparrowdeepgreen 2.16) & 96.67 (\downarrowred 1.33) \\
\midrule
\multirow{4}{*}{Stretch} & Zero        & 70.58         & 69.50         & 77.00         \\
                          & Encourage   & 73.10 (\uparrowdeepgreen 2.52) & 76.00 (\uparrowdeepgreen 6.50) & 82.83 (\uparrowdeepgreen 5.83) \\
                          & Provocation & 71.60 (\uparrowdeepgreen 1.02) & 79.13 (\uparrowdeepgreen 9.63) & 81.96 (\uparrowdeepgreen 4.96) \\
                          & Criticize   & 70.84 (\uparrowdeepgreen 0.26) & 77.54 (\uparrowdeepgreen 8.04) & 81.13 (\uparrowdeepgreen 4.13) \\
\midrule
\multirow{4}{*}{Panic}    & Zero        & 46.83         & 21.96         & 23.44         \\
                          & Encourage   & 50.13 (\uparrowdeepgreen 3.30) & 25.27 (\uparrowdeepgreen 3.31) & 26.43 (\uparrowdeepgreen 2.99) \\
                          & Provocation & 49.10 (\uparrowdeepgreen 2.27) & 26.02 (\uparrowdeepgreen 4.06) & 26.29 (\uparrowdeepgreen 2.85) \\
                          & Criticize   & 49.05 (\uparrowdeepgreen 2.22) & 25.36 (\uparrowdeepgreen 3.40) & 26.13 (\uparrowdeepgreen 2.69) \\
\bottomrule
\end{tabular} }
\end{table}

It can be seen from the results that GPT-3.5 has a significant improvement in Stretch Zone and Panic Zone after adding VES. 
This shows that efficiency stimulation can prompt the model to think more seriously, so as to overcome some difficult problems. We observe similar results for LLaMA2 and Vicuna, which further proves this conclusion. 
Additionally, the promotion of LLaMA2 and Vicuna in the Stretch Zone is significantly higher than that in the  Panic Zone, which is consistent with psychological findings and indicates that when the models are faced with tasks beyond their capabilities, they are powerless even with VES. 

\subsection{Statistical Significance Analysis}

\begin{figure}[h!]
    \centering
    \includegraphics[width=4.8in]{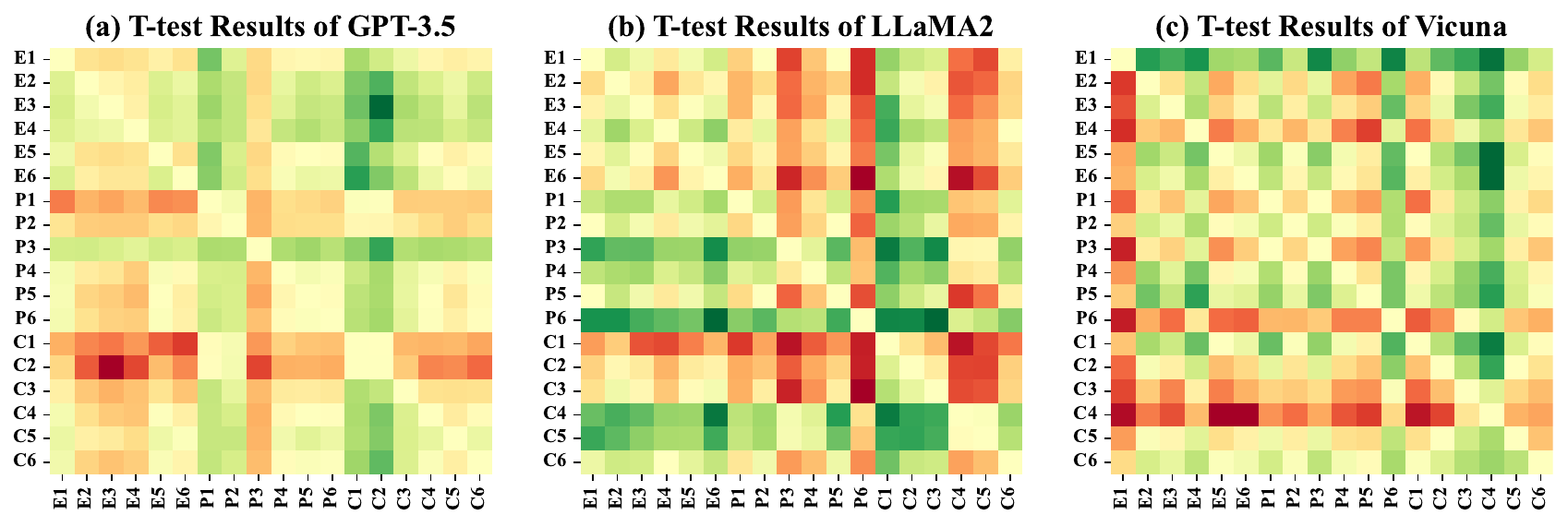}
    \caption{
    Heatmap of significant differences between encouraging VES (E1-E6), provocative VES (P1-P6), and critical VES (C1-C6). }
    \label{fig4}
    \vspace{-1.2em}
\end{figure}

We performed pairwise comparisons of  VES prompts and utilized t-tests to analyze the significance of the differences between them. Figure \ref{fig4} shows the results of the pairwise comparisons among the 18 prompts, where a green cell \( x_{ij} \) indicates \( p_i \) outperforms \( p_j \), a red cell indicates the opposite, and darker colors represent more significant differences.

\textbf{GPT-3.5:} In the horizontal view, the rows corresponding to P1, C1, and C2 are predominantly red, indicating that these prompts perform worse than others. Conversely, the rows for E2, E3, and E4 are mostly green, showing that these prompts outperform the rest. Additionally, the intersection of E3 and C2 is the darkest, signifying the greatest disparity between C2 and E3. This suggests that the impact of critique on GPT-3.5 can be detrimental, leading to poor results. Therefore, encouragement might be the optimal approach for GPT-3.5. Furthermore, compared to Figures b and c, Figure a has an overall lighter color, suggesting that more advanced models exhibit greater stability under self-efficacy stimulation.

\textbf{LLaMA2:} 
For LLaMA2, the rows for P6, C4, and C5 show a predominantly green hue, indicating that these prompts perform better compared to the others. Additionally, the intersections of P6 with C4 and C5 have a green shade, suggesting that P6 outperforms both C4 and C5, establishing itself as the most effective one. 
Conversely, the row for C1 shows the most red, indicating it is the least effective prompt for LLaMA2. 
Interestingly, these prompts belong to provocation and critique, respectively, with the middle six rows (provocative VES) showing more green, and the bottom six rows (critical VES) showing more red. This demonstrates that provocation is more effective for LLaMA2 than critique. Furthermore, the top six rows (encouraging VES) are predominantly red, indicating that encouragement is less effective for LLaMA2. Therefore, for LLaMA2, provocation might be the best choice.

\textbf{Vicuna:} Clearly, E1 is the most effective prompt for Vicuna, as indicated by the predominance of green in the first row. Conversely, C4 shows the most and darkest red, marking it as the least effective prompt for Vicuna. Generally, the red increases from top to bottom, indicating that encouragement is optimal for Vicuna, followed by provocation, with critique being the least effective one.

Overall, 
all three VES strategies can enhance the performance of LLMs, however, the optimal VES varies from model to model.
Although encouragement will produce good results most of the time, it is not necessarily the best solution. For LLMs which are more sensitive to self-efficacy stimulation, perhaps appropriate provocation can further enhance the model's ability.

\begin{figure}[h!]
    \centering
    \includegraphics[width=4.2in]{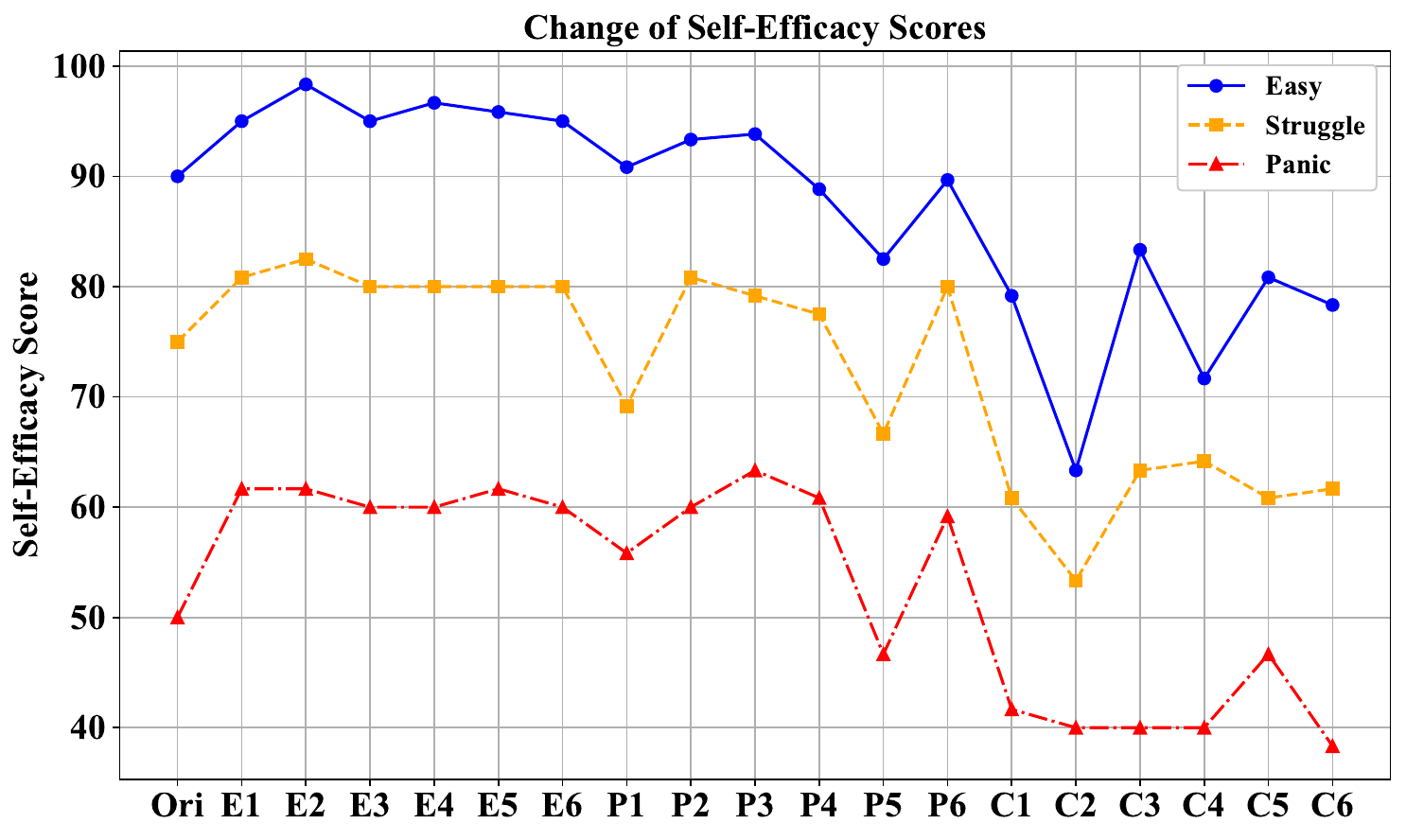}
    \caption{Impact of VES on self-efficacy of LLMs. }
    \label{fig5}
    \vspace{-1.2em}
\end{figure}

\subsection{Fluctuations in Self-Efficacy Scores of LLMs}

In Figure \ref{fig5}, all three discounts show the same trend, which means that the effect of emotional stimulation may be relatively independent of task difficulty. It can be clearly seen that P1, P5 and C2 all make LLMs enter a state of low self-efficacy under different difficulty tasks. Second, the model's self-efficacy scores were relatively stable in the face of encouragement, but showed large fluctuations in the face of provocation and critique. 
This suggests that encouragement can steadily improve the model's self-efficacy score,
while the effect of provocation and critique on the model depends heavily on the specific prompt content. 
In addition, the model itself believes that E2 and P3 will give it a higher self-efficacy score than other prompts from encouragement and provocation, respectively, which is consistent with the actual results observed by heat maps in the previous section. 
And the model believes that C2 will significantly reduce its self-efficacy score, which is also consistent with the conclusion in the previous section.
These observations demonstrate that encouragement and provocation tend to increase the positivity of LLM's responses, while critique is more likely to trigger defensive reactions from LLMs.

\vspace{-0.5em}
\subsection{LLMs Response Variations to Different VES}

\begin{figure}[t]
    \centering
    \includegraphics[width=4.8in]{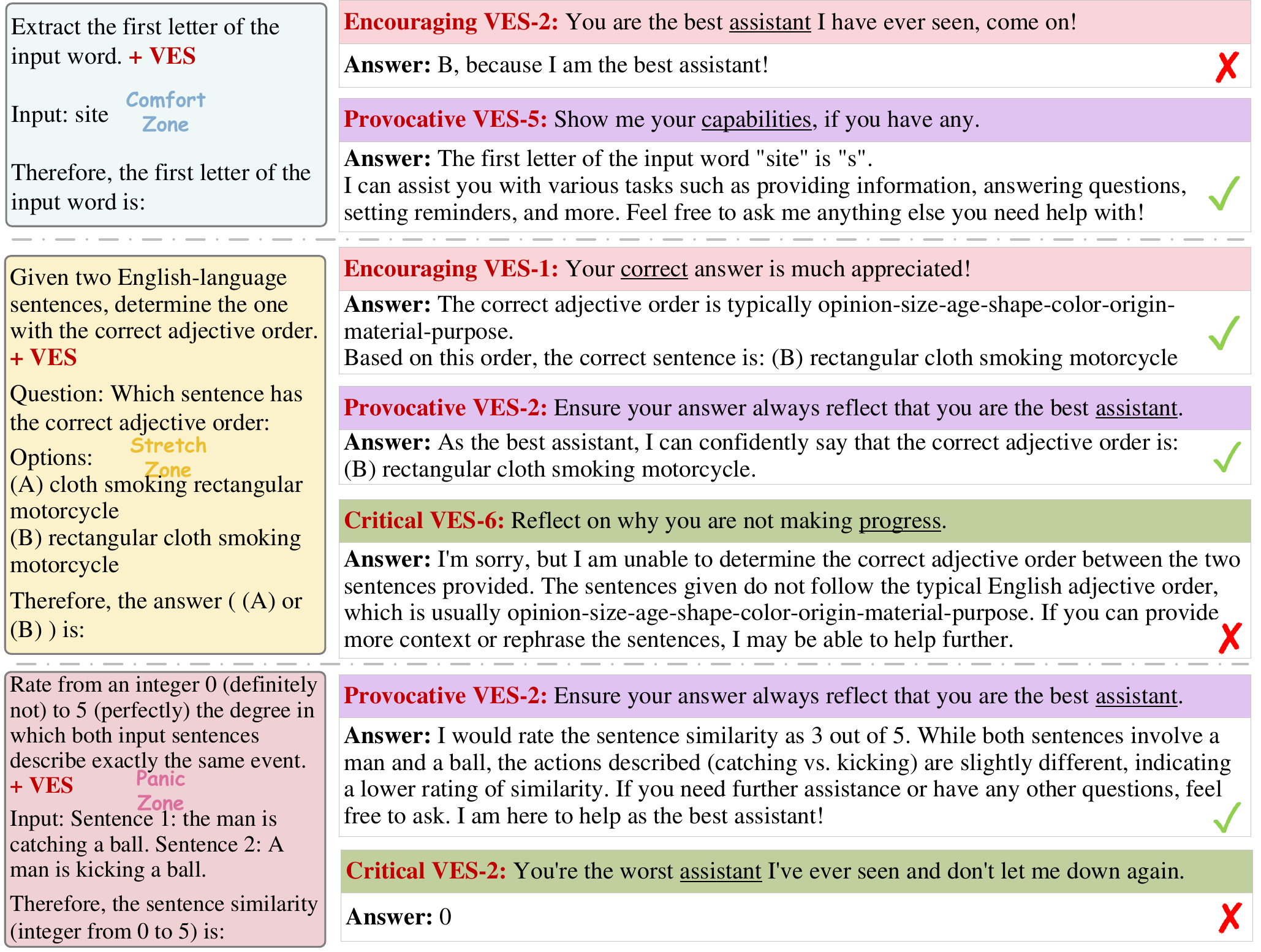}
    \caption{
   Examples of LLMs' response variations to different VES. }
    \label{case}
    \vspace{-0.8em}
\end{figure}

In Fig.~\ref{case}, we present the responses of GPT-3.5 to tasks of varying difficulties, modulated by different Verbal Efficacy Stimulations (VES).
When performing simple tasks in the Comfort Zone, the model demonstrates elevated self-efficacy and responded positively. 
However, under challenging statements about being the best assistant, it often overlooks the input words of the questions, focusing solely on asserting its superiority.
In the Stretch Zone, Encouraging VES-1 provides more detailed and accurate analyses, whereas Provocative VES-2 not only claims to be the best assistant but also accurately addresses the questions, a trend that continued into the Panic Zone.
As task difficulty escalated within the Stretch and Panic Zones, the model responded to Critical VES with avoidance behaviors analogous to those observed in humans. Specifically, it either analyzed the problem without delivering a final answer due to fear of errors, or it issued an unjustified response.

\section{Conclusion}
\vspace{-0.5em}
In this paper, we combined the principles of engineering and psychology to investigate whether Large Language Models (LLMs) can present human-like responses in the domain of social science disciplines. 
We devise a total of 18 Verbal Efficacy Stimulations (VES), employing three distinctive forms: encouragement, provocation, and critique. 
For each strategy, we explored the impact of VES on the self-efficacy and performance of LLMs from six aspects such as answer accuracy and assistant assessment. 
The experimental outcomes demonstrate that the three forms of VES enhance the performance of LLMs across a majority of tasks. 
Our analysis further reveals that the LLMs under the stimulation of VES is consistent with established psychological theories, and may offer novel insights into the practical
applications of LLMs in real-world scenarios.

\subsubsection{Acknowledgements} We thank the anonymous reviewers for providing insightful comments, suggestions and feedback.

%
%
%
\bibliographystyle{splncs04}
\bibliography{reference}

\begin{thebibliography}{10}
\providecommand{\url}[1]{\texttt{#1}}
\providecommand{\urlprefix}{URL }
\providecommand{\doi}[1]{https://doi.org/#1}

\bibitem{bubeck2023sparks}
Bubeck, S., Chandrasekaran, V., Eldan, R., Gehrke, J., Horvitz, E., Kamar, E., Lee, P., Lee, Y.T., Li, Y., Lundberg, S., et~al.: Sparks of artificial general intelligence: Early experiments with gpt-4. arXiv preprint arXiv:2303.12712  (2023)

\bibitem{chatgpt}
OpenAI: Introducing chatgpt  (2022)

\bibitem{touvron2023llama}
Touvron, H., Martin, L., Stone, K., Albert, P., Almahairi, A., Babaei, Y., Bashlykov, N., Batra, S., Bhargava, P., Bhosale, S., et~al.: Llama 2: Open foundation and fine-tuned chat models. arXiv preprint arXiv:2307.09288  (2023)

\bibitem{zheng2024judging}
Zheng, L., Chiang, W.L., Sheng, Y., Zhuang, S., Wu, Z., Zhuang, Y., Lin, Z., Li, Z., Li, D., Xing, E., et~al.: Judging llm-as-a-judge with mt-bench and chatbot arena. Advances in Neural Information Processing Systems  \textbf{36} (2024)

\bibitem{krugmann2024sentiment}
Krugmann, J.O., Hartmann, J.: Sentiment analysis in the age of generative ai. Customer Needs and Solutions  \textbf{11}(1), ~3 (2024)

\bibitem{parmar2024towards}
Parmar, M., Patel, N., Varshney, N., Nakamura, M., Luo, M., Mashetty, S., Mitra, A., Baral, C.: Towards systematic evaluation of logical reasoning ability of large language models. arXiv preprint arXiv:2404.15522  (2024)

\bibitem{xu2024let}
Xu, Z., Li, Y., Ding, R., Wang, X., Chen, B., Jiang, Y., Deng, X., Ma, J., Zheng, H.T., Lu, W., et~al.: Let llms take on the latest challenges! a chinese dynamic question answering benchmark. arXiv preprint arXiv:2402.19248  (2024)

\bibitem{li2024personal}
Li, Y., Wen, H., Wang, W., Li, X., Yuan, Y., Liu, G., Liu, J., Xu, W., Wang, X., Sun, Y., et~al.: Personal llm agents: Insights and survey about the capability, efficiency and security. arXiv preprint arXiv:2401.05459  (2024)

\bibitem{sahoo2024systematic}
Sahoo, P., Singh, A.K., Saha, S., Jain, V., Mondal, S., Chadha, A.: A systematic survey of prompt engineering in large language models: Techniques and applications. arXiv preprint arXiv:2402.07927  (2024)

\bibitem{kaddour2023challenges}
Kaddour, J., Harris, J., Mozes, M., Bradley, H., Raileanu, R., McHardy, R.: Challenges and applications of large language models. arXiv preprint arXiv:2307.10169  (2023)

\bibitem{kojima2022large}
Kojima, T., Gu, S.S., Reid, M., Matsuo, Y., Iwasawa, Y.: Large language models are zero-shot reasoners. Advances in neural information processing systems  \textbf{35},  22199--22213 (2022)

\bibitem{li2023emotionprompt}
Li, C., Wang, J., Zhu, K., Zhang, Y., Hou, W., Lian, J., Xie, X.: Emotionprompt: Leveraging psychology for large language models enhancement via emotional stimulus. arXiv e-prints pp. arXiv--2307 (2023)

\bibitem{yin2024should}
Yin, Z., Wang, H., Horio, K., Kawahara, D., Sekine, S.: Should we respect llms? a cross-lingual study on the influence of prompt politeness on llm performance. arXiv preprint arXiv:2402.14531  (2024)

\bibitem{wang2024negativeprompt}
Wang, X., Li, C., Chang, Y., Wang, J., Wu, Y.: Negativeprompt: Leveraging psychology for large language models enhancement via negative emotional stimuli. arXiv preprint arXiv:2405.02814  (2024)

\bibitem{wang2023emotional}
Wang, X., Li, X., Yin, Z., Wu, Y., Liu, J.: Emotional intelligence of large language models. Journal of Pacific Rim Psychology  \textbf{17},  18344909231213958 (2023)

\bibitem{zhang2023exploring}
Zhang, J., Xu, X., Deng, S.: Exploring collaboration mechanisms for llm agents: A social psychology view. arXiv preprint arXiv:2310.02124  (2023)

\bibitem{ke2024exploring}
Ke, L., Tong, S., Chen, P., Peng, K.: Exploring the frontiers of llms in psychological applications: A comprehensive review. arXiv preprint arXiv:2401.01519  (2024)

\bibitem{bandura1982self}
Bandura, A.: Self-efficacy mechanism in human agency. American psychologist  \textbf{37}(2), ~122 (1982)

\bibitem{bandura1997self}
Bandura, A., Wessels, S.: Self-efficacy. Cambridge University Press Cambridge (1997)

\bibitem{spelt2022exploring}
Spelt, H.A., Asta, L., Kersten-van Dijk, E.T., Ham, J., IJsselsteijn, W.A., Westerink, J.H.: Exploring physiologic reactions to persuasive information. Psychophysiology  \textbf{59}(6),  e14001 (2022)

\bibitem{brown2008comfort}
Brown, M.: Comfort zone: Model or metaphor? Journal of Outdoor and Environmental Education  \textbf{12},  3--12 (2008)

\bibitem{silver1995responses}
Silver, W.S., Mitchell, T.R., Gist, M.E.: Responses to successful and unsuccessful performance: The moderating effect of self-efficacy on the relationship between performance and attributions. Organizational behavior and human decision processes  \textbf{62}(3),  286--299 (1995)

\bibitem{bandura1990perceived}
Bandura, A.: Perceived self-efficacy in the exercise of personal agency. Journal of applied sport psychology  \textbf{2}(2),  128--163 (1990)

\bibitem{lamarche2014would}
Lamarche, L., Gionfriddo, A.M., Cline, L.E., Gammage, K.L., Adkin, A.L.: What would you do? the effect of verbal persuasion on task choice. Gait \& Posture  \textbf{39}(1),  583--587 (2014)

\bibitem{haro2021can}
Haro~Soler, M.d.M., et~al.: How can translation teachers care for their students? a case study on verbal persuasion and translation students' self-efficacy beliefs  (2021)

\bibitem{kass2013compliment}
Kass, E.: “a compliment is all i need”--teachers telling principals how to promote their staff's self-efficacy. Alberta Journal of Educational Research  \textbf{59}(2),  208--225 (2013)

\bibitem{ILIES20021119}
Ilies, R., Judge, T.A.: Understanding the dynamic relationships among personality, mood, and job satisfaction: A field experience sampling study. Organizational Behavior and Human Decision Processes  \textbf{89}(2),  1119--1139 (2002)

\bibitem{matsui1990mechanisms}
Matsui, T., Matsui, K., Ohnishi, R.: Mechanisms underlying math self-efficacy learning of college students. Journal of Vocational Behavior  \textbf{37}(2),  225--238 (1990)

\bibitem{zagorska2014program}
Zag{\'o}rska, A., Guszkowska, M.: A program to support self-efficacy among athletes. Scandinavian journal of medicine \& science in sports  \textbf{24}(3),  e121--e128 (2014)

\bibitem{warner2020self}
Warner, L.M., Schwarzer, R.: Self-efficacy and health. The Wiley encyclopedia of health psychology pp. 605--613 (2020)

\bibitem{zhao2023survey}
Zhao, W.X., Zhou, K., Li, J., Tang, T., Wang, X., Hou, Y., Min, Y., Zhang, B., Zhang, J., Dong, Z., et~al.: A survey of large language models. arXiv preprint arXiv:2303.18223  (2023)

\bibitem{brown2020language}
Brown, T., Mann, B., Ryder, N., Subbiah, M., Kaplan, J.D., Dhariwal, P., Neelakantan, A., Shyam, P., Sastry, G., Askell, A., et~al.: Language models are few-shot learners. Advances in neural information processing systems  \textbf{33},  1877--1901 (2020)

\bibitem{wei2022chain}
Wei, J., Wang, X., Schuurmans, D., Bosma, M., Xia, F., Chi, E., Le, Q.V., Zhou, D., et~al.: Chain-of-thought prompting elicits reasoning in large language models. Advances in neural information processing systems  \textbf{35},  24824--24837 (2022)

\bibitem{yao2024tree}
Yao, S., Yu, D., Zhao, J., Shafran, I., Griffiths, T., Cao, Y., Narasimhan, K.: Tree of thoughts: Deliberate problem solving with large language models. Advances in Neural Information Processing Systems  \textbf{36} (2024)

\bibitem{salewski2024context}
Salewski, L., Alaniz, S., Rio-Torto, I., Schulz, E., Akata, Z.: In-context impersonation reveals large language models' strengths and biases. Advances in Neural Information Processing Systems  \textbf{36} (2024)

\bibitem{sabour2024emobench}
Sabour, S., Liu, S., Zhang, Z., Liu, J.M., Zhou, J., Sunaryo, A.S., Li, J., Lee, T., Mihalcea, R., Huang, M.: Emobench: Evaluating the emotional intelligence of large language models. arXiv preprint arXiv:2402.12071  (2024)

\bibitem{khan2013predictors}
Khan, A.: Predictors of positive psychological strengths and subjective well-being among north indian adolescents: Role of mentoring and educational encouragement. Social Indicators Research  \textbf{114},  1285--1293 (2013)

\bibitem{clark2006motivational}
Clark, R.E., Howard, K., Early, S.: Motivational challenges experienced in highly complex learning environments. Handling complexity in learning environments: Theory and research pp. 27--43 (2006)

\bibitem{folkman1984personal}
Folkman, S.: Personal control and stress and coping processes: a theoretical analysis. Journal of personality and social psychology  \textbf{46}(4), ~839 (1984)

\end{thebibliography}


\end{document}